\documentclass[10pt,twocolumn,letterpaper]{article}

\usepackage{iccv}
\usepackage{times}
\usepackage{epsfig}
\usepackage{graphicx}
\usepackage{amsmath}
\usepackage{amssymb}

\usepackage{color}
\usepackage{calc}
\usepackage{algorithm,algpseudocode}
\usepackage{booktabs}
\usepackage{multirow}

\usepackage{makecell}
\usepackage{float}
\usepackage{subfig}

\usepackage{color}

\newcommand{\gray}[1]{\textcolor[rgb]{0.7,0.7,0.7}{#1}}

\DeclareMathOperator{\similarity}{sim}
\DeclareMathOperator{\sg}{sg}

\usepackage[pagebackref=true,breaklinks=true,letterpaper=true,colorlinks,bookmarks=false]{hyperref}

\iccvfinalcopy

\ificcvfinal\fi

\begin{document}

\title{Asymmetric Patch Sampling for Contrastive Learning}

\author{Chengchao Shen\textsuperscript{1}, Jianzhong Chen\textsuperscript{1}, Shu Wang\textsuperscript{1}, 
Hulin Kuang\textsuperscript{1}, Jin Liu\textsuperscript{1}, Jianxin Wang\textsuperscript{1}\\
\textsuperscript{1}Central South University\\
{\tt\small \{scc.cs,cjz\_csu,wangshu.dr,hulinkuang,liujin06\}@csu.edu.cn, jxwang@mail.csu.edu.cn}
}

\maketitle
\ificcvfinal\thispagestyle{empty}\fi

\begin{abstract}
   Asymmetric appearance between positive pair effectively reduces the risk of representation degradation in contrastive learning.
   However, there are still a mass of appearance similarities between positive pair constructed by the existing methods, which inhibits the further representation improvement.

   In this paper, we propose a novel asymmetric patch sampling strategy for contrastive learning, to further boost the appearance asymmetry for better representations. 
   Specifically, dual patch sampling strategies are applied to the given image, to obtain asymmetric positive pairs. 
   First, sparse patch sampling is conducted to obtain the first view, which reduces spatial redundancy of image and allows a more asymmetric view.
   Second, a selective patch sampling is proposed to construct another view with large appearance discrepancy relative to the first one.

   Due to the inappreciable appearance similarity between positive pair, the trained model is encouraged to capture the similarity on semantics, instead of low-level ones. 
   Experimental results demonstrate that our proposed method significantly outperforms the existing self-supervised methods on both ImageNet-1K and CIFAR dataset, e.g., 2.5\% finetune accuracy improvement on CIFAR100. 
   Furthermore, our method achieves state-of-the-art performance on downstream tasks, object detection and instance segmentation on COCO.
   Additionally, compared to other self-supervised methods, our method is more efficient on both memory and computation during training. 
   The source code is available at \url{https://github.com/visresearch/aps}.
\end{abstract}

\section{Introduction}
{
   \begin{figure}[t]
      \centering
      \includegraphics[width=0.9\linewidth]{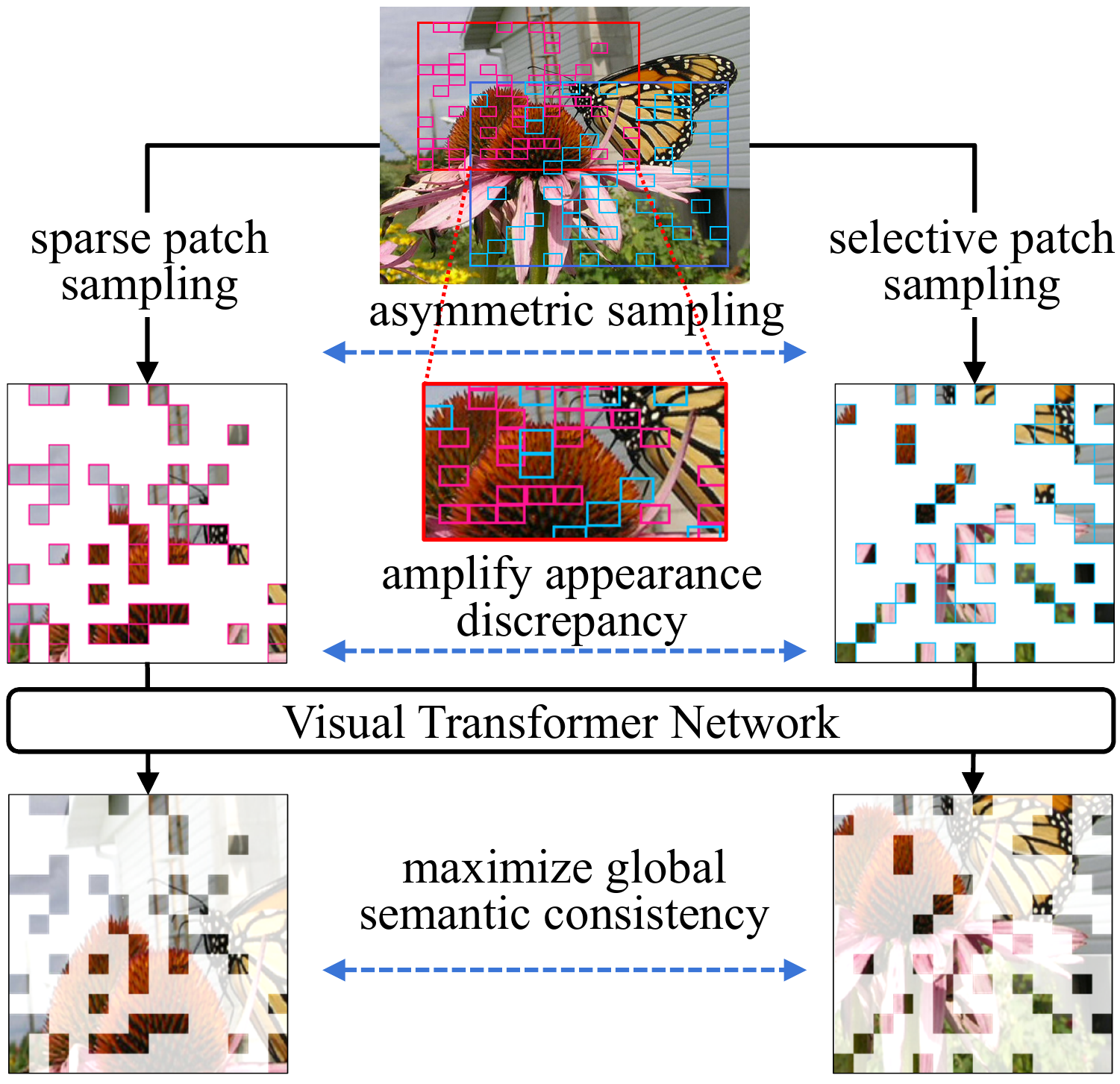}
      \caption{Asymmetric patch sampling strategy to construct positive pairs with large appearance discrepancy, which effectively reduces the appearance similarity but without changing image semantics. 
      By contrasting positive pairs with inappreciable appearance similarity, the trained model is encouraged to capture the similarity from the perspective of image semantics.}
      \label{fig:motivation}
   \end{figure}

   In recent years, massive breakthroughs are achieved in self-supervised/unsupervised learning field. 
   Based on the difference of pretext tasks, the popular branches contain contrastive learning (CL)~\cite{wu2018unsupervised,chen2020simple,he2020momentum,grill2020bootstrap,chen2021exploring,zbontar2021barlow,wang2022importance} and masked image modeling (MIM)~\cite{xie2022simmim,dong2021peco,he2022masked,gao2022convmae,chen2022context}.

   For contrastive learning task, the trained model is required to discriminate different views of the same images from other images, named instance discrimination~\cite{wu2018unsupervised}. 
   To learn semantic representations of images, contrastive learning methods introduce a series of asymmetric designs, such as data augmentation\cite{chen2020simple}, to increase the appearance discrepancy between positive pairs, but without changing image semantics.
   In this way, the trained model is encouraged to understand the semantics in images, instead of some trivial features.
   Therefore, plausible asymmetric designs are significantly important for the performance of contrastive learning.
   However, due to image overlap between positive pair, there is still a mass of appearance similarity in the existing contrastive learning methods, which degrades the representations.

   Different from contrastive learning, MIM task follows the idea of masked language modeling task (MLM)~\cite{radford2018improving,radford2019language,brown2020language,devlin2019bert} in natural language processing (NLP), where partially masked data are fed into the model to predict the invisible part of data in an auto-encoding manner. 
   Due to the heavy spatial redundancy of image, the highly random masked images in MIM task can still effectively retain the semantics of the original images~\cite{he2022masked}, which achieves very promising performance in self-supervised learning.
   However, with the similar semantics, the raw pixels or their tokens have a large fluctuation on appearance, leading to non-unique solutions to invisible patch reconstruction from the random masked images, especially when the masked ratio is large. 
   The existing MIM methods attempt to map the highly masked image into a fixed target, which inevitably introduces large fitting error, even if the prediction is a plausible solution for the given input.
   We call it as \emph{non-unique target issue}, which substantially limits the flexibility of the MIM models.

   Inspired by the above observations, we propose a novel asymmetric patch sampling strategy, to introduce more asymmetry for contrastive learning and alleviate the non-unique target issue suffered by the existing MIM methods at the same time.
   Specifically, the improvement is two fold. 
   First, to improve the semantics of contrastive pretext task, the proposed sampling strategy constructs positive pairs, where the two views are essentially semantic consistent but with inappreciable similarity on appearance.
   Second, compared to MIM methods, we replace the reconstruction objective with contrastive one, which provides more flexible targets for training.

   As shown in Figure~\ref{fig:motivation}, our proposed method respectively adopts two different patch sampling strategies for two views of positive pair, named \emph{asymmetric patch sampling strategy} (\emph{APS}). 
   For the first view, we conduct sparse patch sampling~\cite{he2022masked} to obtain highly sparse patch sequences, which only contain small portion of patches from the original image, e.g., 25\%. 
   This operation aims to reduce spatial redundancy of image and encourage the Visual Transformer (ViT) network to model long distance dependency among patches. 
   For the second view, we conduct a selective patch sampling, where the patches not appearing in the first view are preferred to be sampled.
   In this way, the appearance of the sampled views are essentially different from each other. 
   To minimizing the contrastive objective, the ViT model is required to understand the semantics of images. 
   Furthermore, we formally analyze the asymmetry of the proposed method.
   To improve training stability of contrastive learning, we further propose an adaptive gradient clip operation.

   In summary, our main contribution is an asymmetric patch sampling strategy to construct efficient positive pairs for contrastive learning, which allows two views to represent the same object with essentially different appearances. 
   The trained model is hard to fit the contrastive objective by only simply comparing low-level features. 
   Thus, it is encouraged to extract semantic representations from the asymmetric positive pairs. 
   Experimental results demonstrate that the proposed APS achieves excellent performance on unsupervised representation learning. 
   Specifically, for both ViT-Small/16 and ViT-Base/16, our method significantly outperforms the previous best results on ImageNet-1K. 
   When pretrained on small datasets, CIFAR10 and CIFAR100, our method respectively surpasses the previous state-of-the-art method by $1.2\%$ and $2.5\%$ using ViT-Tiny/2 backbone.
   Furthermore, our method also achieve state-of-the-art performance on downstream tasks, object detection and instance detection on COCO.
}

\begin{figure*}[t]
   \centering
   \includegraphics[width=\linewidth]{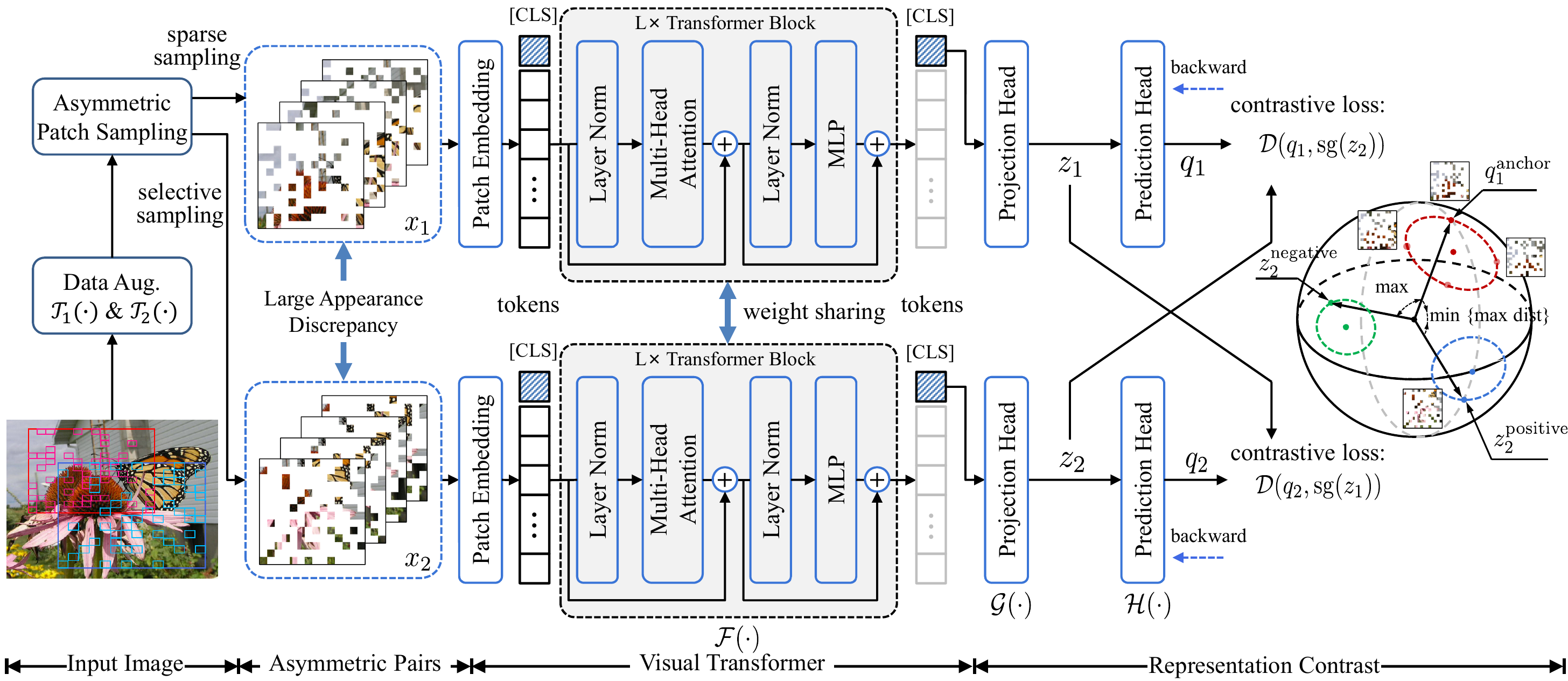}
   \caption{The overview of asymmetric patch sampling for contrastive learning.
   Dual asymmetric patch sampling strategies, sparse sampling and selective sampling, are conducted on the input image, to construct positive pairs with large appearance discrepancy.
   The visual transformer encoder is trained to minimize distance between positive pairs obtained by asymmetric sampling (hard samples) and meanwhile maximize distance between negative pairs.
   }
   \label{fig:overview}
\end{figure*}

\section{Related Work}
{
   \subsection{Masked Image Modeling}
   {
      Following MLM task~\cite{radford2018improving,radford2019language,brown2020language,devlin2019bert} in NLP, masked image modeling is proposed as a novel pretext task for self-supervised learning in computer vision. 
      In this task, the masked raw pixels~\cite{he2022masked,fang2022corrupted,xie2022simmim} or their tokens~\cite{bao2022beit,dong2021peco,wei2022masked} are used as the targets for model training. 
      However, with the similar semantics, the raw pixels or their tokens have a large fluctuation on appearance, which may provide not so stable supervision signal. 
      Moreover, the solutions of the given masked image are generally not the unique one, especially when the masked ratio is high, which may cause large prediction error on a plausible prediction and effect the flexibility of model learning. 

      To this end, feature prediction based methods are proposed to alleviate the above issue. 
      iBOT~\cite{zhou2022ibot} performs self-distillation to recover invisible patch tokens from visible ones. 
      CAE~\cite{chen2022context} adopts the representation alignment between visible patches and invisible patches before decoder in MIM task. 
      SIM~\cite{tao2022siamese} introduces siamese network structure to align semantics between the tokens from different augmented views by mask operation. 
      In spite of the encouraging results achieved, these methods still heavily rely on the unstable targets for pixel/token reconstruction.

      In contrast, the proposed method models masked images by learning instance similarity between significantly different views sampled by asymmetric sampling strategy, which provides more flexible and stable targets for self-supervised learning. 
      
   }

   \subsection{Contrastive Learning}
   {
      Contrastive learning conducts instance classification by maximizing representation similarity under different distortions of a sample (positive pair) and minimizing the one of different samples (negative pair), to learn invariant representation of data under different distortions~\cite{wu2018unsupervised,chen2020simple,he2020momentum,grill2020bootstrap,chen2021exploring,zbontar2021barlow,wang2022importance}. 
      To avoid trivial solutions and learn valid representation, asymmetric designs play a vital role in contrastive learning, which introduce a series of variances on low-level features but without changing the semantics of images~\cite{zbontar2021barlow,wang2022importance}. 
      The most important asymmetric design is a series of data augmentation techniques applied on positive pair, which is the most widely adopted by popular contrastive learning methods~\cite{chen2020simple,chen2020improved,grill2020bootstrap,caron2020unsupervised,chen2020big}. 
      For example, color jitter, gray scale and solarize operation significantly change the color of images in positive pairs, so the model in the contrastive setting is required to capture the color-invariant representations. 
      Then random crop operation introduces variance on object parts and scale, which further removes the dependency on object parts and scale for model. 
      Therefore, the model is trained to recognize objects using semantic features, instead of trivial ones. 
      The asymmetric designs are also introduced into network architectures, such as prediction head~\cite{grill2020bootstrap,chen2021exploring,chen2021empirical} and momentum encoder~\cite{he2020momentum,chen2020improved,grill2020bootstrap,chen2021empirical,caron2021emerging}, which disturb the representation of positive pairs.

      In this work, we introduce a novel effective asymmetric operation, asymmetric patch sampling strategy.
      It constructs a series of significant different positive pairs, each of which contains quite little appearance similarity with their distorted version. 
      To minimizing the contrastive objective, the model is encouraged to learn more semantic and useful representation. 

   }

   \subsection{Hard Sample Mining}
   {
      Hard sampling mining is widely adopted in object detection~\cite{shrivastava2016training,lin2017focal}, to alleviate the extreme foreground-background class imbalance of dataset. 
      OHEM~\cite{shrivastava2016training} selects the ROI with large loss to update the model during training. 
      Focal loss~\cite{lin2017focal} dynamically scales cross entropy loss for training samples, which encourages the model to focus on hard samples. 

      These methods tap hard samples according to the loss value during training for object detection. 
      However, they also boost the potential negative effect caused by mislabeled samples. 
      In contrast, our method directly constructs hard positive pairs by sampling different patch combinations from the same objects, which obtains informative samples with fewer potential mislabeled ones. 

   }
   
}

\section{Method}
{

   \subsection{Overview}
   {
      In the setting of contrastive learning, it's believed that positive pairs with large appearance discrepancy can effectively regularize the representation learning. 
      To further improve appearance discrepancy, we adopt sparse patch sampling strategy as masked image modeling methods, in which only small portion of patches are sampled to construct positive pairs. 
      In this way, two views of positive pair have fewer overlapping patches and effectively improve the asymmetry on low-level features. 

      As shown in Figure~\ref{fig:overview}, the input image $x$ is randomly cropped into two sub-images $x_1^{\prime}$ and $x_2^{\prime}$, and then processed by data augmentation $\mathcal{T}_1(\cdot)$ and $\mathcal{T}_2(\cdot)$ to obtain the positive pair for contrastive learning, which can be represented as 
      \begin{align}\label{eq:aug}
         [x_1^{\prime}, B_1] = \mathcal{T}_1(x), \\
         [x_2^{\prime}, B_2] = \mathcal{T}_2(x), \nonumber
      \end{align}
      where $B_i (i = 1,2)$ denotes the bounding box produced by random image crop.
      
      To further reduce the appearance similarity on patches, we conduct asymmetric sampling strategy $\mathcal{A}(\cdot)$, including sparse sampling and selective sampling, to lower the overlapping probability between the two target views as follows,
      \begin{align}\label{eq:asym}
         [x_1, x_2] = \mathcal{A}(x_1^{\prime}, x_2^{\prime}, B_1, B_2; s_1, s_2),
      \end{align}
      where $s_i (i = 1, 2)$ denotes the patch sampling ratio of $x_i^{\prime}$, e.g., $s_1 = s_2 = 0.25$. 

      Then, combined with projection head $\mathcal{G}(\cdot)$, the positive pair $[x_1, x_2]$ is respectively fed into visual transformer $\mathcal{F}(\cdot)$ to obtain the representation $z_i = \mathcal{G}(\mathcal{F}(x_i))$.
      Additionally, prediction head module $\mathcal{H}(\cdot)$ is adopted, to increase the asymmetry between the representations of positive pair. 
      The output can be written as $q_i = \mathcal{H}(z_i)$.

      Finally, we conduct contrastive learning between projection representation $z_i$ and prediction representation $q_i$, to learn appearance-invariant representations from positive pair with extremely asymmetric appearance. 

   }

   \begin{figure}[t]
      \centering
      \includegraphics[width=0.8\linewidth]{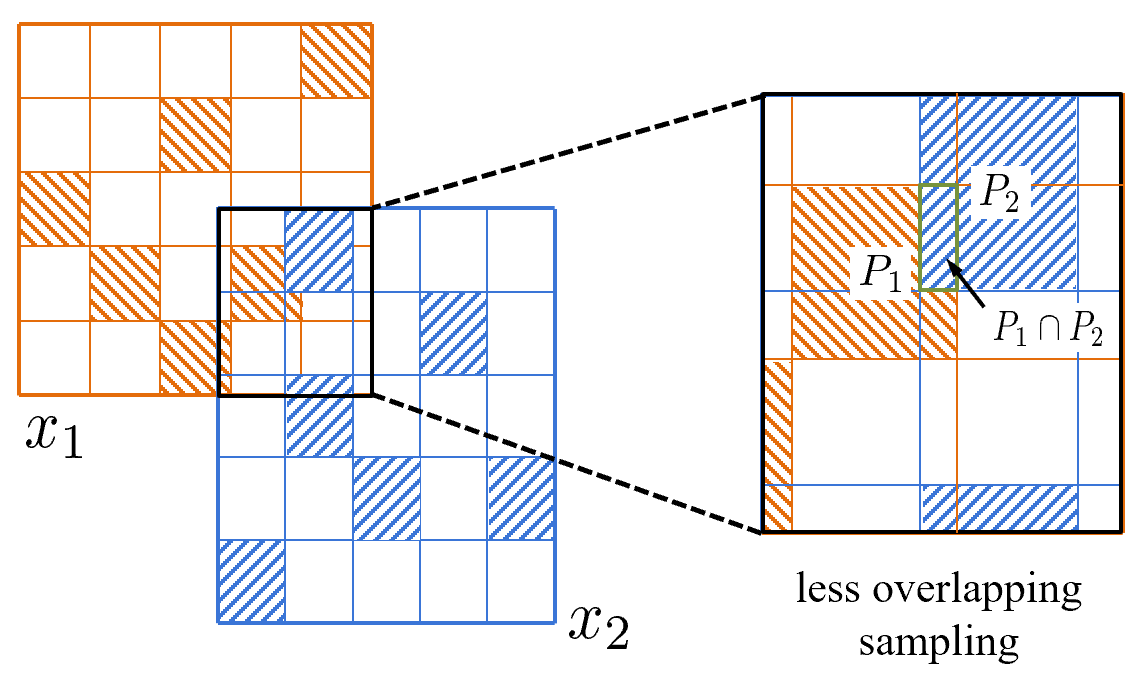}
      \caption{The overlap ratio computation between two randomly cropped views. 
      }
      \label{fig:asymmetric_patch_sample}
   \end{figure}

   \subsection{Asymmetric Patch Sampling}
   {
      As shown in Figure~\ref{fig:asymmetric_patch_sample}, the overlap ratio between patches from $x_1^\prime$ and $x_2^\prime$ can be computed as follows, 
      \begin{align}\label{eq:overlap}
         r_{\rm overlap} = \frac{\mathcal{S}(P_1 \cap P_2)}{\mathcal{S}(P_2)},
      \end{align}
      where $P_i (i = 1,2)$ denotes the sampled patch in view $i$ and   $\mathcal{S}(\cdot)$ denotes the area of the given patch. 
      To reduce the sampling probability of overlapping patches, we propose a selective patch sampling method, whose sampling probability density $p_{\rm sel}$ is computed by
      \begin{align}\label{eq:sampling}
         p_{\rm sel} = (\gamma + 1) \cdot s_1 \cdot (1 - r_{\rm overlap})^\gamma,
      \end{align}
      where $\gamma$ is a hyper-parameter to tune the sensitivity of sampling. 
      As shown in Figure~\ref{fig:sampling}, the larger $\gamma$ is, the smaller possibility of the sampled overlapping patches is. 
      This selective sampling method and sparse sampling form the asymmetric sampling strategy. 

      Since sparse sampling strategy uniformly samples patch from the first view $x_1$, the probability of patches sampling in the overlapping area $x_1 \cap x_2$ is also $s_1$ as the one in $x_1$.
      Hence, the sampling probability density is required to meet the following equation, 
      \begin{align}\label{eq:condition}
         \int_{0}^{1} p_{\rm sel} {\rm d}r_{\rm overlap} = s_1,
      \end{align}
      which guarantees that the total probability meets the ratio of our proposed sampling strategy in $x_1 \cap x_2$\footnote{More details can be found in the supplementary material.}. 
      In other words, all areas under the curves in Figure~\ref{fig:sampling} equal to $s_1$.
      
   }

   \begin{figure}[ht]
      \centering
      \includegraphics[width=\linewidth]{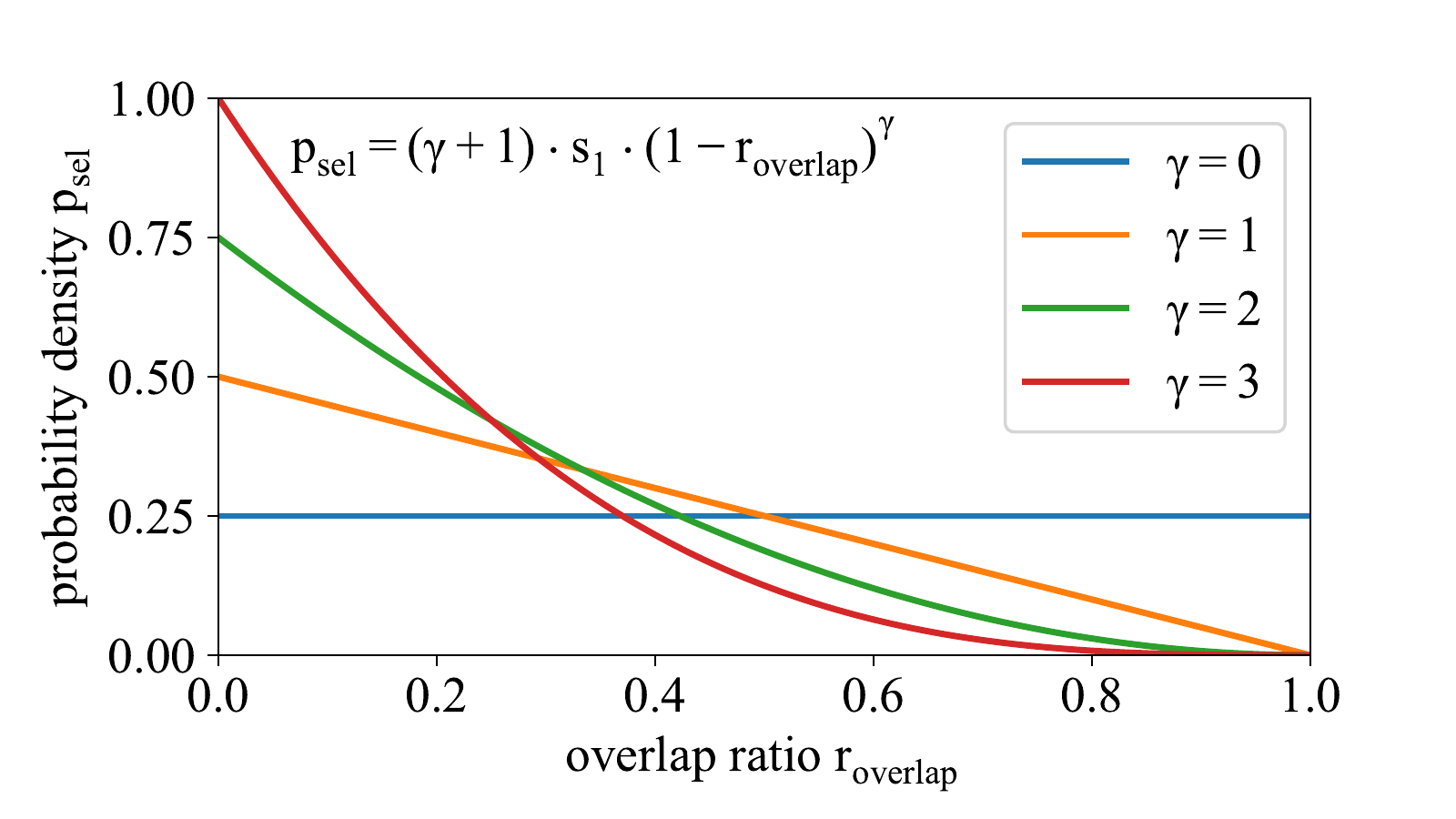}
      \caption{Asymmetric sampling curve.
      The higher the overlap ratio is, the smaller the patch sampling probability is.
      }
      \label{fig:sampling}
   \end{figure}

   \subsection{Asymmetry Analysis}
   {
      In this section, we formally analyze the asymmetry of patch sampling strategy, to quantify the appearance discrepancy between positive pairs. 
      For convenience, we focus on the spatial asymmetry, which is simply defined as the non-overlap ratio between the images of positive pair.
      With consideration of sampling randomness, we measure the spatial asymmetry by the expectation of non-overlap ratio between positive pairs, where non-overlap ratio is defined as $r_{\rm non} = 1 - r_{\rm overlap}$ and $r_{\rm overlap}$ denotes the overlap between images from positive pair. 
      The expectation of non-overlap ratio $\mathbb{E}_{\rm non}$ can be obtained from the expectation of overlap ratio $\mathbb{E}_{\rm overlap}$ by $\mathbb{E}_{\rm non} = 1 - \mathbb{E}_{\rm overlap}$. 
      For brevity, we focus on the analysis of $\mathbb{E}_{\rm overlap}$ as follows.

      As shown in Figure~\ref{fig:asymmetry} (a), a naive patch sampling strategy uniformly samples patches from the same image crop to construct positive pairs for contrastive learning. 
      The sampling probability of each patch in crop $x_1^{\prime}$ and $x_2^{\prime}$ can be regarded as their sampling ratios, $s_1$ and $s_2$, respectively. 
      This patch sampling strategy subjects to bernoulli distribution, where each patch in the crop takes overlap probability of $s_1 \cdot s_2$ and non-overlap probability of $(1 - s_1 \cdot s_2)$.
      Therefore, the expectation of overlap ratio between positive pair by the naive path sampling strategy can be computed as
      \begin{align}\label{eq:naive-overlap}
         \mathbb{E}_{\rm overlap}^{\rm naive} = 1 \cdot s_1 \cdot s_2 + 0 \cdot (1 - s_1 \cdot s_2) = s_1 \cdot s_2.
      \end{align}
      This sparse patch sampling strategy can effectively reduce the overlap between positive pair. 
      For example, when $s_1 = s_2 = 0.25$, the expectation of ratio $\mathbb{E}_{\rm overlap}^{\rm naive} = 0.0625$, which means only $6.25\%$ region is overlapped in the two views of positive pairs on average and significantly improves asymmetry. 

      \begin{figure}[t]
         \centering
         \includegraphics[width=\linewidth]{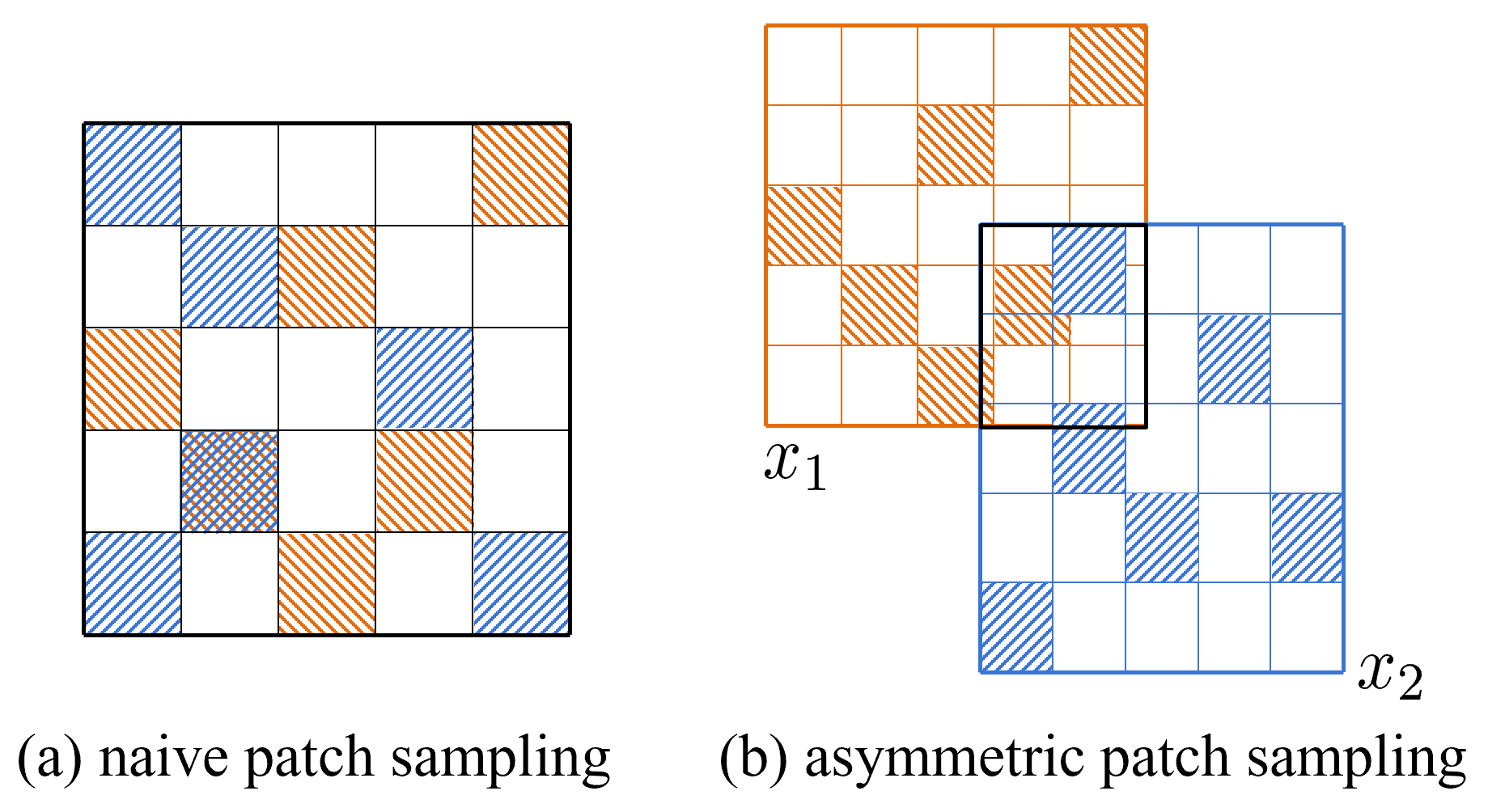}
         \caption{Asymmetry analysis.
         Two alternative patch sampling strategies. 
         (a) randomly sample patches from the same cropped image from origin image; 
         (b) combine random crop operation and random patch sampling one.
         }
         \label{fig:asymmetry}
      \end{figure}

      To further improve the asymmetry between positive pair, we combine the random image crop operation and the asymmetric patch sampling strategy as Figure~\ref{fig:asymmetry} (b). 
      So the overlap ratio between the patches from $x_1$ and $x_2$ varies from 0 to 1 ($r_{\rm overlap} \in [0, 1]$), instead of discrete values, $0$ and $1$, as naive patch sampling strategy, which allows introducing more asymmetry. 
      Specifically, we conduct asymmetric patch sampling using Eq.~\ref{eq:sampling}. 
      As naive patch sampling, the overlap probability density of asymmetric sampling can be computed by $p_{\rm overlap} = p_{\rm sel} \cdot s_2$.
      So the expectation of overlap ratio between positive pair can be obtained by
      \begin{align}\label{eq:asym-overlap}
         \mathbb{E}_{\rm overlap}^{\rm sel} = \int_0^1{p_{\rm overlap} \cdot r_{\rm overlap} {\rm d} r_{\rm overlap}}
         = \frac{s_1 \cdot s_2}{\gamma + 2}.
      \end{align}
      In other words, the overlap ratio of asymmetric patch sampling strategy is $\frac{1}{\gamma + 2}$ of the one by naive patch sampling strategy, which effectively improves the asymmetry between positive pair. 
      For example, when $\gamma = 3$, the expectation of overlap ratio is only $20 \%$ of the naive one according to Eq.~\ref{eq:naive-overlap} and Eq.~\ref{eq:asym-overlap}.

   }

   \subsection{Optimization}
   {
      In this section, we use for reference loss function in~\cite{grill2020bootstrap,chen2021exploring,chen2020simple,chen2021empirical} to form our contrastive objective as:
      \begin{align}\label{eq:asym-loss}
         \mathcal{L}_{\rm contrast} = \tau \cdot \left[\mathcal{D}(q_1, \sg(z_2)) + \mathcal{D}(q_2, \sg(z_1)) \right], \\
         \mathcal{D}(q, z) = - \sum_{i=1}^{N} \log \frac{\exp(\similarity(q^{(i)}, z^{(i)}) / \tau)}{\sum_{j=1}^{N}\exp(\similarity(q^{(i)}, z^{(j)}) / \tau )}, \nonumber \\
         \similarity(q, z) = \frac{q^{\rm T} \cdot z}{\Vert q \Vert \cdot \Vert z \Vert}, \nonumber
      \end{align}
      where $\tau$ and $N$ respectively denote temperature parameter and batch size, $q^{(i)}$ and $z^{(i)}$ respectively denote the representation $q$ and $z$ of $i$-th sample in mini-batch, $\sg(\cdot)$ denotes stop gradient operation\footnote{Temperature $\tau$ in $\mathcal{L}_{\rm contrast}$ is used to simplify the learning rate tuning under different temperature values. More details can be found in the supplementary material}. 
      Compared to~\cite{grill2020bootstrap,chen2021exploring,chen2020simple,chen2021empirical}, the main difference of our objective function is the combination of the normalized temperature-scaled cross entropy loss and stop gradient operation. 

      As clarified in MoCo-v3~\cite{chen2021empirical}, contrastive learning methods for vision transformer tend to suffer an unstable optimization issue, which significantly hurts the performance. 
      To stabilize the training, gradient clip strategy is a popular technique, which scales $\ell_2$-norm of the excessive gradients to the given maximal norm. 
      However, this clip strategy fails to adapt to the case, where the gradients progressively decay but sometimes abnormal fluctuation during contrastive training.
      To solve this issue, we set an adaptive threshold for step $t$ according to the exponential moving average of gradient $\mathcal{G}_t$ as follows: 
      \begin{align}\label{eq:grad-ema}
         \mathcal{G}_t = m \cdot \mathcal{G}_{t - 1} + (1 - m) \cdot g_t,
      \end{align}
      where $m \in [0, 1)$ is a momentum coefficient and $g_t$ denotes the gradients with respect to model parameters at step $t$. 
      When $ \Vert g_{t} \Vert > \alpha \cdot \Vert \mathcal{G}_{t-1} \Vert$, the gradient $g_{t}$ is scaled by the norm of threshold $\mathcal{G}_{t-1}$: 
      \begin{align}\label{eq:grad-scale}
         \hat{g}_t = g_t \cdot \frac{\Vert \mathcal{G}_{t-1} \Vert}{ \Vert g_t \Vert + \epsilon },
      \end{align}
      where $\epsilon$ is set to $10^{-8}$ for better numerical stability, 
      to adjust the amplitude of gradients into a reasonable range and stabilize the training of contrastive learning. 

   }

}

\section{Experiments}
{
   In this section, we conduct experiments to evaluate the effectiveness of our proposed method. 
   First, we give the experiment settings, including model backbone, optimization hyperparameters and other details. 
   Then, we compare our method with state-of-the-art methods on ImageNet-1K and CIFAR.
   Finally, we further analyze the effectiveness of each component in our method by ablation study.

   \begin{table*}[ht]
   \centering
   {
      \begin{tabular}{lccccccc}
         \toprule[1pt]
         \textbf{Method} & \textbf{Backbone} & \textbf{Add. Module} & \textbf{\#Views} & \textbf{Supervision} & \textbf{\#Epochs} & \textbf{Finetune (\%)} \\ 
         \hline
         MoCo v3~\cite{chen2021empirical} & ViT-S/16       & pred. head             & 2     & contrastive rep.                   & 300             & 81.4  \\
         \hline
         MAE~\cite{he2022masked} & ViT-S/16       & ViT decoder   & 1     & invisible pixel                             & 1600           & 79.5 \\
         \hline
         CAE~\cite{chen2022context} & ViT-S/16    & \makecell{DALL-E \& \\ ViT decoder}   & 2     & \makecell{invisible \\ visual token} & 300            & 81.8 \\
         \hline
         BEiT~\cite{bao2022beit} & ViT-S/16       & \makecell{DALL-E \& \\ ViT decoder} & 2     & \makecell{invisible \\ visual token}        & 300            & $81.7$  \\
         iBOT~\cite{zhou2022ibot}& ViT-S/16       & proj. head            & 2     & self distillation  & 800 & 81.8\\
         \bottomrule[1pt]
         \textbf{APS (ours)} & ViT-S/16           & pred. head             & 2     & contrastive rep.                   & 300            & \textbf{82.1} \\
         \textbf{APS (ours)} & ViT-S/16           & pred. head             & 2     & contrastive rep.                   & 800            & \textbf{82.6} \\
         \bottomrule[1pt]
         \gray{supervised}~\cite{dosovitskiy2021vit}      & \gray{ViT-B/16}       & \gray{/}             & \gray{1}     & \gray{images from JFT}                              & \gray{/}               & \gray{79.9} \\
         \hline
         MoCo v3~\cite{chen2021empirical} & ViT-B/16       & pred. head             & 2     & contrastive rep.                   & 300             & 83.0 \\
         \hline
         DINO~\cite{caron2021emerging} & ViT-B/16       & pred. head             & 12    & contrastive rep.                   & 400             & 82.8 \\
         \hline
         MAE~\cite{he2022masked} & ViT-B/16       & ViT decoder   & 1     & invisible pixel                              & 1600             & 83.6 \\
         \hline
         CAE~\cite{chen2022context} & ViT-B/16       & \makecell{DALL-E \& \\ ViT decoder}   & 2     & \makecell{invisible \\ visual token} & 800            & 83.6 \\
         \hline
         BEiT~\cite{bao2022beit} & ViT-B/16       & \makecell{DALL-E \& \\ ViT decoder} & 2     & \makecell{invisible \\ visual token}        & 300              & 83.2 \\
         \hline
         MixMIM~\cite{liu2022mixmim} & ViT-B/16       & ViT decoder   & 2     & invisible pixel                              & 300              & 83.2 \\
         \hline
         SimMIM~\cite{xie2022simmim} & ViT-B/16       & pred. head    & 1     & invisible pixel                              & 800              & 83.8 \\
         \hline
         MaskFeat~\cite{wei2022masked} & ViT-B/16     & pred. head    & 1     & HOG feature                                  & 1600             & 84.0 \\
         \hline
         iBOT~\cite{zhou2022ibot}      & ViT-B/16     & proj. head    & 14     & self distillation                           & 800              & 83.8 \\
         \bottomrule[1pt]
         \textbf{APS (ours)} & ViT-B/16               & pred. head    & 2     & contrastive rep.                   & 300              & \textbf{84.2} \\
         \textbf{APS (ours)} & ViT-B/16               & pred. head    & 2     & contrastive rep.                   & 800              & \textbf{84.7} \\
         \bottomrule[1pt]

      \end{tabular}
   }
   
   \caption{Performance comparison on ImageNet-1K dataset. 
   ``pred. head'' and ``proj. head'' respectively denote prediction and projection head.
   ``contrastive rep.'' denotes contrastive representation.
   ``NA'' denotes the result is not available in the original paper.
   }
   \label{table:imagenet}
   \end{table*}

   \begin{table*}[ht]
      \centering
      \resizebox{\linewidth}{!}
      {
         \begin{tabular}{l|c|c|c|cc|cc}
            \hline
            \multicolumn{1}{l|}{} & \multicolumn{1}{l|}{} & \multicolumn{1}{l|}{} & \multicolumn{1}{l|}{} & \multicolumn{2}{c|}{\textbf{CIFAR10}} & \multicolumn{2}{c}{\textbf{CIFAR100}} \\
            \textbf{Method} & \textbf{Backbone} & \textbf{Memory (G)}  & \textbf{\#FLOPs (G)}  & \textbf{\#Epochs} & \textbf{Accuracy (\%)} & \textbf{\#Epochs} & \textbf{Accuracy (\%)} \\ 
            \hline
            \gray{supervised}  & \gray{ViT-T/2} & \gray{/} & \gray{/} &   \gray{300}       & \gray{89.6}  & \gray{300} &  \gray{68.1} \\
            MoCo-v3~\cite{chen2021empirical}   & ViT-T/2  & 40.0 & 21.9 &    1600  &  96.3    & 1600 & 81.4 \\
            MAE~\cite{he2022masked}            & ViT-T/2  & 13.3 & 4.8 &    1600  &  96.1    & 1600 & 76.3 \\
            SimMIM~\cite{xie2022simmim}        & ViT-T/2  & 28.6 & 6.0 &    1600  &  95.3    & 1600 & 79.1 \\
            iBOT~\cite{zhou2022ibot}           & ViT-T/2  & 58.5 & 59.6 &    1600  &  96.3    & 1600 & 80.1 \\
            \textbf{APS (ours)}                & ViT-T/2  & \textbf{9.0} & \textbf{4.2} &    1600 & \textbf{97.4} & 1600 & \textbf{83.9} \\
            \textbf{APS (ours)}                & ViT-T/2  & \textbf{9.0} & \textbf{4.2} &    3200 & \textbf{97.4} & 3200 & \textbf{83.8} \\
            \hline
            \gray{supervised} & \gray{ViT-S/2} & \gray{/} & \gray{/} & \gray{300} & \gray{90.3} & \gray{300} &  \gray{68.9} \\
            MoCo-v3~\cite{chen2021empirical}   & ViT-S/2  & 80.0 & 87.6 &   1600  &  97.3    & 1600 & 83.1 \\
            MAE~\cite{he2022masked}            & ViT-S/2  & 25.5 & 19.2 &   1600  &  97.0    & 1600 & 80.0 \\
            SimMIM~\cite{xie2022simmim}        & ViT-S/2  & 57.9 & 24.6 &   1600  &  97.1    & 1600 & 83.2 \\
            iBOT~\cite{zhou2022ibot}           & ViT-S/2  & 105.3 & 238.5 &   1600  &  97.0    & 1600 & 82.8 \\
            \textbf{APS (ours)}                & ViT-S/2  & \textbf{16.2} & \textbf{16.2} &   1600  &  \textbf{98.0} & 1600 & \textbf{84.8} \\
            \textbf{APS (ours)}                & ViT-S/2  & \textbf{16.2} & \textbf{16.2} &   3200  &  \textbf{98.1} & 3200 & \textbf{85.6} \\
            \hline
            \gray{IN pretrained}~\cite{touvron2021training} & \gray{ViT-B/16}  &  \gray{/} &  \gray{/} & \gray{300} & \gray{98.1} & \gray{300} & \gray{87.1} \\
            \gray{supervised} & \gray{ViT-B/2} & \gray{/} & \gray{/} & \gray{300} & \gray{92.7} & \gray{300} &  \gray{72.6} \\
            MoCo-v3~\cite{chen2021empirical}   & ViT-B/2  & 159.3 & 348.2 &   1600  &  97.2    & 1600 & 82.6 \\
            MAE~\cite{he2022masked}            & ViT-B/2  & 50.8  & 76.2  &   1600  &  94.7    & 1600 & 72.3 \\
            SimMIM~\cite{xie2022simmim}        & ViT-B/2  & 115.3 & 97.6  &   1600  &  97.1  & 1600 & 82.8 \\
            \textbf{APS (ours)}                & ViT-B/2  & 33.1 & 66.2   &   1600  &  \textbf{98.2} & 1600 & \textbf{85.1} \\
            \textbf{APS (ours)}                & ViT-B/2  & 33.1 & 66.2   &   3200  &  \textbf{98.2} & 3200 & \textbf{86.1} \\
            \hline
         \end{tabular}
      }
      \caption{Performance comparison on CIFAR dataset using finetune evaluation protocol.
      ``Memory'' denotes GPU memory consumed during self-supervised training, where batch size is set to 512 and automatic mixed precision is applied.
      ``\#FLOPs'' denotes the number of floating-point operations per iteration during training, where batch size is set to 1.
      ``IN pretrained'' denotes the model pretrained on ImageNet-1K using supervised cross entropy loss and then finetuned on CIFAR with $224 \times 224$ size.
      }
      \label{table:cifar}
   \end{table*}

   \subsection{Experimental Settings}
   {
      \subsubsection{Data Processing}
      For both ImageNet-1K~\cite{ILSVRC15} and CIFAR~\cite{krizhevsky2009learning}, we sample $25\%$ patches from the given images, namely sampling ratio $s_1 = s_2 = 0.25$. 
      The sizes of input image are $224 \times 224$ and $32 \times 32$ for ImageNet-1K and CIFAR, respectively.
      Additionally, more data augmentations applied on ImageNet-1K and CIFAR are clarified in the supplementary material.

      \subsubsection{Network Architecture}
      For ImageNet-1K, we adopt ViT-Small/16 and ViT-Base/16~\cite{dosovitskiy2021vit} (respectively denoted as \emph{ViT-S/16} and \emph{ViT-B/16}) as the backbone, respectively. 
      Following MoCo-v3~\cite{chen2021empirical}, projection head and prediction head module are added and the detail can be found in the supplementary material. 
      During pretraining, we adopt fixed sine-cosine position embedding and random initialized patch embedding, which are combined with adaptive gradient operation to improve training stability. 
      The momentum encoder is also applied, whose ema coefficient varies from $0.99$ to $1.0$ as cosine scheduler.

      For CIFAR10 and CIFAR100, we conduct experiments on ViT-Tiny/2 and ViT-Small/2 (respectively denoted as \emph{ViT-T/2} and \emph{ViT-S/2}), whose structures are depicted in the supplementary material.
      Different from experiments on ImageNet-1K, both position and patch embedding are learnable, without training instability issue.
      Additionally, momentum encoder is not adopted on CIFAR.

      \subsubsection{Optimization}
      For loss function, the temperature $\tau$ is set to $0.1$ on both ImageNet-1K and CIFAR.
      For ImageNet-1K, we use AdamW optimizer~\cite{loshchilov2019decoupled} with batch size $4096$, learning rate $1.28 \times 10^{-3}$, momentum $0.9$ and weight decay $0.1$. 
      During pretraining, we conduct learning rate warmup for $20$ epochs and then follow a cosine learning rate decay schedule for the rest $780$ epochs. 
      To further stabilize the training, adaptive gradient clip operation is implemented on each transformer block, where $m = 0.4$ and $\alpha = 1.05$.
      For patch sampling, we set $\gamma = 3$ to increase the appearance discrepancy between positive pair.

      For CIFAR10 and CIFAR100, we use AdamW optimizer with batch size $512$, learning rate $1 \times 10^{-3}$, momentum $0.9$ and weight decay $0.05$. The model is trained by $1,600$ epochs, where the first $20$ epochs for warmup.
      Additionally, gradient clip operation is not conducted on the model for CIFAR.

      \subsubsection{Evaluation}
      Linear probing has been a popular evaluation protocol for contrastive learning. 
      However, it fails to evaluate the quality of non-linear features~\cite{bao2022beit,he2022masked,noroozi2016unsupervised}. 
      For our proposed method, the input image fed into network is highly sparse instead of complete one during training, which is significantly different from the input samples during evaluation. 
      Hence, we adopt finetune protocol to evaluate the representation of our model, instead of linear probing one.
      By default, all pretrained models are finetuned for 100 epochs.

   }

   \subsection{Image Classification}
   {

      \subsubsection{Results on ImageNet-1K}
      {
         As show in Table~\ref{table:imagenet}, our proposed method achieves $82.6\%$ finetune accuracy on ViT-B/16, which significantly outperforms the previous SOTA method iBOT by $0.8\%$ finetune accuracy. 
         Moreover, our proposed method achieves $84.7\%$ finetune accuracy on ViT-B/16, which outperforms the previous SOTA method iBOT by $0.9\%$ finetune accuracy.
         We believe that our method tends to extract more semantic representation by modeling representation similarity between asymmetric positive pairs.
         Moreover, our method does not require additional tokenizer, such as DALL-E~\cite{ramesh2021zero} to provide visual tokens as supervision signal, which is more efficient and concise.
         Compared to previous state-of-the-arts, MaskFeat and iBOT, our method requires shorter training schedule but achieves better performance.
      }

      \subsubsection{Results on CIFAR}
      {
         As shown in Table~\ref{table:cifar}, our method with ViT-T/2 achieves $97.4\%$ and $83.9\%$ finetune accuracy on CIFAR10 and CIFAR100, respectively, surpassing the previous state-of-the-art method, iBOT by $1.1\%$ and $3.8\%$ on CIFAR10 and CIFAR100, respectively. 
         For ViT-S/2 backbone, our method consistently outperforms the previous best finetune accuracy by $1.0\%$ on CIFAR10 and $2.0\%$ on CIFAR100.
         For ViT-B/2 backbone, our proposed method achieves $98.2\%$ and $86.1\%$ accuracy on CIFAR10 and CIFAR100, respectively, which significantly reduces the performance gap with the model pretrained on large-scale dataset: ImageNet-1K in a fully supervised manner. 
         Especially on CIFAR10, our method (input image size: $32 \times 32$) evens outperform the ImageNet-1K pretrained one using ViT-B/16 (input image size: $224 \times 224$). 
         Overall, our method achieves the best performance on CIFAR dataset without extra data, even outperforms the model pretrained by large-scale dataset: ImageNet-1K.
         
         Moreover, our method consumes the least memory during training, only about $\frac{1}{6}$ memory of iBOT and $\frac{1}{5}$ memory of MoCo-v3 on both ViT-T/2 and ViT-S/2.
         Therefore, our method is more friendly to hardware than other self-supervised methods, which sheds some light on more training-efficient self-supervised learning method. 
         Meanwhile, our method is the most computation-efficient self-supervised method, only about $\frac{1}{15}$ computation of iBOT and $\frac{1}{5}$ computation of MoCo-v3, thus requires less training time.

      }
      
   }

   \subsection{Transfer Learning on Downstream Tasks}

   To further evaluate the transferability of our method, we conduct transfer learning experiments on downstream tasks: object detection and instance segmentation on COCO~\cite{lin2014microsoft} by Mask RCNN~\cite{he2017mask} framework. 
   As shown in Table~\ref{table:coco}, our proposed APS achieves 51.8 box AP and 46.2 mask AP on objection detection and instance segmentation, respectively.
   Compared to previous SOTA method iBOT, our APS achieves 0.6 box AP and 2.0 mask AP improvement, which illustrates the effectiveness of our method.

   \begin{table}[ht]
      \centering
      {
         \begin{tabular}{lccc}
            \toprule[1pt]
            \textbf{Method}                  & \textbf{\#Pre-Epochs} & \textbf{AP}$^{\bf box}$ & \textbf{AP}$^{\bf mask}$\\ 
            \hline
            MoCo v3~\cite{chen2021empirical} & 300                   & 47.9       & 42.7 \\
            \hline
            MAE~\cite{he2022masked}          & 1600                  & 50.3       & 44.9 \\
            \hline
            CAE~\cite{chen2022context}       & 300                   & 48         & 42.3 \\
            \hline
            iBOT~\cite{zhou2022ibot}         & 1600                  & 51.2       & 44.2 \\
            \hline
            \textbf{APS (ours)}              & 800                   & \textbf{51.8} & \textbf{46.2} \\
            
            \bottomrule[1pt]

         \end{tabular}
      }
      \caption{Object detection and instance segmentation using pretrained ViT-B/16 model on COCO. 
      }
      \label{table:coco}
   \end{table}

   \subsection{Ablation Study}
   {
      In this section, we implement ablation study to validate the efficiency of modules in our method. 
      For convenience, the experiments are mainly conducted on CIFAR dataset using ViT-T and partially on ImageNet dataset using ViT-S.

      \subsubsection{The Effect of Patch Sampling Ratio}
      To evaluate the effectiveness of patch sampling ratio, we compare the performance using different ratios on CIAFAR10 and CIFAR100. 
      As shown in Table~\ref{table:sampling-ratio}, our method respectively reports $97.4\%$ and $83.9\%$ accuracy on CIFAR10 and CIFAR100, when sampling ratio $s = 0.25$.
      This experimental result on CIFAR100 outperforms the one with $s = 1.0$ by $1.6\%$, which demonstrates that patch sampling strategy can effectively improve the representation quality of contrastive learning.
      In contrast, the experiment using extremely sparse sampling ratio $s = 0.15$, leads to a $1.9\%$ degradation on CIFAR100, which significantly damages the performance.
      We believe that reasonable sparse sampling strategy applied on images can effectively regularize the contrastive representation learning.

      \begin{table}[ht]
         \centering
         {
            \begin{tabular}{ccc}
               \toprule[1pt]
               \textbf{sampling ratio ($s$)} & \textbf{CIFAR10 (\%)} & \textbf{CIFAR100 (\%)}  \\ 
               \bottomrule[1pt]
               1.0            & 96.4  & 82.3 \\
               0.75           & 96.9  & 81.9 \\
               0.5            & 97.0  & 83.2 \\
               0.25           & \textbf{97.4} & \textbf{83.9} \\
               0.15           & 96.7  & 82 \\
               \bottomrule[1pt]
            \end{tabular}
         }
         \caption{Performance comparison on different patch sampling ratios.}
         \label{table:sampling-ratio}
      \end{table}

      \subsubsection{The Effect of Sampling Power}
      To analyze the effect of sapling power $\gamma$ in Eq.~\ref{eq:sampling}, we conduct experiments on CIFAR10 and CIFAR100 using different values. 
      As shown in Table~\ref{table:sampling-power}, the model achieves the best performance on both CIFAR10 and CIFAR100 when $\gamma = 3.0$.
      Especially on CIFAR100, the model with $\gamma = 3.0$ outperforms the one of $\gamma=0.0$ by $2.6\%$. 
      It demonstrates that the model can benefit from a reasonable sampling power $\gamma$, thus is inclined to extract semantic representations from asymmetric positive pairs.
      However, when $\gamma=4.0$, it does not achieve better performance than the one of $\gamma = 1.0$ or $3.0$. 
      The possible reason is that excessive punishment on highly overlapping patches may reduce the chance of learning some important representations from the neglected patches.

      For CIFAR10, the performance fluctuation of our method is pretty small under different $\gamma$ values. 
      This can be in part explained by that, due to less category diversity of CIFAR10, the trained model can not apparently benefit from asymmetric patch sampling strategy. 

      \begin{table}[ht]
         \centering
         \resizebox{\linewidth}{!}
         {
            \begin{tabular}{ccc}
               \toprule[1pt]
               \textbf{sampling power ($\gamma$)} & \textbf{CIFAR10 (\%)} & \textbf{CIFAR100 (\%)}  \\ 
               \bottomrule[1pt]
               0.0            & 97.0   & 81.3   \\
               1.0            & 97.0   & 82.0   \\
               2.0            & 97.3   & 82.5   \\
               3.0            & \textbf{97.4} & \textbf{83.9} \\
               4.0            & 97.0   & 82.8   \\
               \bottomrule[1pt]
            \end{tabular}
         }
         \caption{Performance comparison on different asymmetric sampling power values.}
         \label{table:sampling-power}
      \end{table}

      \subsubsection{The Effect of Sampling View Number}
      Since only small portion of patches are used during the training, we reuse the rest patches of loaded data to improve the data throughout and training efficiency. 
      Furthermore, we compare the performance when different numbers of sampling views are used to training.
      As shown in Table~\ref{table:sampling-view}, the experimental results demonstrate that using all sampling views can significantly improve the performance of model.
      Especially on CIFAR100, 1.5\% accuracy improvement is achieved. 
      Additionally, we find that more sampling views can efficiently improve training stability during experiments.

      \begin{table}[ht]
         \centering
         {
            \begin{tabular}{ccc}
               \toprule[1pt]
               \textbf{\#sampling} & \textbf{CIFAR10 (\%)} & \textbf{CIFAR100 (\%)}  \\ 
               \bottomrule[1pt]
               1 & 96.8   & 82.4   \\
               2 & 96.9   & 82.5   \\
               3 & 97.2   & 83.3   \\
               4 & \textbf{97.4} & \textbf{83.9} \\
               \bottomrule[1pt]
            \end{tabular}
         }
         \caption{Performance comparison on different number of sampling views.
         In this case, the sampling ratio is $0.25$ and up to $4$ views can be used for training.}
         \label{table:sampling-view}
      \end{table}

      \subsubsection{The Effect of Adaptive Gradient Clip}
      To investigate the effectiveness of adaptive gradient clip operation, we conduct experiments on small datasets, CIFAR using ViT-T/2 and large-scale dataset, ImageNet-1K using ViT-S/16 for 300 epochs. 
      The experimental results are shown in Table~\ref{table:grad-clip}, demonstrating that the adaptive gradient clip achieves 1.2\% accuracy improvement on ImageNet-1K.
      On the contrary, applying adaptive gradient clip operation on CIFAR degrades the performance.
      Due to larger sample diversity of ImageNet-1K, the dramatic sample variation among consecutive sample batches is easy to cause instability of model training. 
      The adaptive gradient clip can effectively improve the training stability of model on large-scale dataset.
      Additionally, we find that clip momentum $m = 0.4$ can significantly improve the model performance on ImageNet-1K. 

      \begin{table}[ht]
         \centering
         \resizebox{\linewidth}{!}
         {
            \begin{tabular}{cccc}
               \toprule[1pt]
               \textbf{grad. clip} ($m$) & \textbf{CIFAR10 (\%)} & \textbf{CIFAR100 (\%)} & \textbf{ImageNet-1K (\%)}\\ 
               \bottomrule[1pt]
               0   & \textbf{97.4} & \textbf{83.9} & 80.9\\
               0.2 & 97.0 & 82.8 & 81.6\\
               0.4 & 97.0 & 82.1 & \textbf{82.1} \\
               0.8 & 96.8 & 81.9 & 81.8 \\
               \bottomrule[1pt]
            \end{tabular}
         }
         \caption{Performance comparison on different gradient clip momentum coefficients. }
         \label{table:grad-clip}
      \end{table}
   }
}

\section{Conclusion}
In this paper, we propose a novel asymmetric patch sampling strategy for contrastive learning. 
This strategy significantly improves the appearance discrepancy between positive pairs and the hardness of pretext task for self-supervised learning. 
Due to fewer clues to similarity on low-level feature between positive pairs, the model is encouraged to learn semantic representations. 
Then, we formally analyze the asymmetry metric of our method and compare it with the baseline. 
Afterwards, we give the optimization objective of our model.
Finally, we propose a novel adaptive gradient clip operation to stabilize the training of our model. 
Experimental results demonstrate that the proposed method is superior to the previous state-of-the-art on both ImageNet-1K, CIFAR and COCO, while consuming less memory and computation. 
For future work, we plan to explore more effective and efficient asymmetric designs to boost the performance of contrastive learning.

{\small
\bibliographystyle{ieee_fullname}
\bibliography{paper.bbl}
}

\clearpage

\appendix

\section{Network Structures}

As shown in Table~\ref{table:backbone}, we give the details of backbones used in our experiments. 
For ImageNet-1K with input size $224 \times 224$, we adopt standard vision transformer architectures, ViT-Small and ViT-Base, where the patch size for tokenization is $16 \times 16$. 
For CIFAR10 and CIFAR100 with input size $32 \times 32$, we modify the patch size of standard vision transformer architectures from $16 \times 16$ to $2 \times 2$, to adapt the small input images. 
We also introduce a more lightweight vision transformer architecture, ViT-Tiny, which only has half head number and half token dimension of ViT-Small.

\begin{table*}[ht]
\centering
{
   \begin{tabular}{ccccccc}
      \toprule[1pt]
      \textbf{Dataset} & \textbf{Network} & \textbf{Patch Size} & \textbf{\#Blocks} & \textbf{\#Heads} & \textbf{Token Dim} & \textbf{\#Params (M)} \\ 
      \toprule[1pt]
      ImageNet-1K & ViT-Small & 16 & 12 & 6   & 384 & 21.6 \\ 
                  & ViT-Base  & 16 & 12 & 12  & 768 & 85.7 \\ 
      \hline
      CIFAR       & ViT-Tiny  & 2  & 12 & 3   & 192 & 5.4  \\ 
                  & ViT-Small & 2  & 12 & 6   & 384 & 21.3 \\ 
                  & ViT-Base  & 2  & 12 & 12  & 768 & 85.1 \\ 
      \bottomrule[1pt]

   \end{tabular}
}

\caption{The structure of visual transformer backbones. 
``\#Blocks'' denotes the number of standard transformer blocks in backbone. 
``Token Dim'' denotes the dimension of visual token vector.
}
\label{table:backbone}
\end{table*}

\begin{table*}[ht]
   \centering
   {
      \begin{tabular}{c|c|l|l}
         \hline
         \textbf{Dataset} & \textbf{Layer} & \makecell[c]{\textbf{Projection Head}} & \makecell[c]{\textbf{Prediction Head}} \\
         \hline
         ImageNet-1K & 1 & Linear (4096) + BN + ReLU & Linear (4096) + BN + ReLU\\
                     & 2 & Linear (4096) + BN + ReLU & Linear (256) + BN\textsuperscript{*} \\
                     & 3 & Linear (256) + BN\textsuperscript{*}              & \\
         \hline
         CIFAR       & 1 & Linear (512) + BN + ReLU  & Linear (512) + BN + ReLU\\
                     & 2 & Linear (512) + BN + ReLU  & Linear (512) + BN + ReLU\\
                     & 3 & Linear (128) + BN\textsuperscript{*}              & Linear (128) + BN\textsuperscript{*}\\
         \hline
      \end{tabular}
   }
   \caption{The structure of projection and prediction heads. 
   ``Linear ($m$)'' denotes linear layer with output size $m$.
   ``BN'' and ``ReLU'' denote batch normalization and rectify linear unit operation, respectively. 
   ``BN\textsuperscript{*}'' denotes batch normalization without learnable parameters.
   }
   \label{table:head}
\end{table*}

As shown in Table~\ref{table:head}, we further present the structure of projection and prediction head adopted during self-supervised pretraining. 
For ImageNet-1K, there are 3 linear layers in projection head and 2 linear layers in prediction heads.
The first two linear layers are followed by batch normalization and rectify linear unit in turn, and the output sizes of them are both 4096. 
The last linear of both heads are followed by only batch normalization, and output sizes of them are 256.
For CIFAR10 and CIFAR100, the configurations of projection and prediction head are similar to the ones of ImageNet-1K.
The main differences can be summarized as follows.
First, the output sizes of projection and prediction are both modified to 128.
Second, the sizes of hidden units are both modified to 512.

\section{Data Augmentations}

As shown in Table~\ref{table:augmentation}, we describe the parameters of data augmentations used during self-supervised pretraining. 
For ImageNet-1K, 6 data augmentation techniques are applied to the input images, including random crop and resize, horizontal flip, color jittering, gray scale, Gaussian blurring, as well as solarization. 
There are two differences between the augmentations for the construction of positive pair.
First, the probability of Gaussian blurring is 1.0 in augmentation $\mathcal{T}_1(\cdot)$, but 0.1 in augmentation $\mathcal{T}_2(\cdot)$.
Second, solarization is only used in augmentation $\mathcal{T}_2(\cdot)$.
For CIFAR10 and CIFAR100, due to small image size, Gaussian blurring is not used.
Additionally, solarization is also not adopted.

\begin{table*}[ht]
   \centering
   \resizebox{\linewidth}{!}
   {
      \begin{tabular}{l|l|cc|cc}
         \hline
         \multicolumn{1}{l|}{} & \multicolumn{1}{l|}{} & \multicolumn{2}{c|}{\textbf{ImageNet-1K}} & \multicolumn{2}{c}{\textbf{CIFAR}} \\
         \makecell[c]{\textbf{Augmentation}} & \makecell[c]{\textbf{Parameter}} & \textbf{Aug.} $\mathcal{T}_1(\cdot)$ & \textbf{Aug.} $\mathcal{T}_2(\cdot)$ & \textbf{Aug.} $\mathcal{T}_1(\cdot)$ & \textbf{Aug.} $\mathcal{T}_2(\cdot)$ \\ 
         \hline
         random crop and resize   & area of the crop                    & [0.08, 1.0] & [0.08, 1.0] & [0.15, 1.0] & [0.15, 1.0] \\
                                  & aspect ratio of the crop            & [$\frac{3}{4}$, $\frac{4}{3}$] & [$\frac{3}{4}$, $\frac{4}{3}$] & [$\frac{3}{4}$, $\frac{4}{3}$] & [$\frac{3}{4}$, $\frac{4}{3}$] \\
         \hline
         random horizontal flip   & horizontal flip probability         & 0.5 & 0.5 & 0.5 & 0.5 \\
         \hline
         random color jittering   & color jittering probability         & 0.8 & 0.8 & 0.8 & 0.8 \\
                                  & max brightness adjustment intensity & 0.4 & 0.4 & 0.4 & 0.4 \\
                                  & max contrast adjustment intensity   & 0.4 & 0.4 & 0.4 & 0.4 \\
                                  & max saturation adjustment intensity & 0.2 & 0.2 & 0.4 & 0.4 \\
                                  & max hue adjustment intensity        & 0.1 & 0.1 & 0.1 & 0.1 \\
         \hline
         random gray scale        & color dropping probability          & 0.2 & 0.2 & 0.2 & 0.2 \\
         \hline
         random Gaussian blurring & Gaussian blurring probability       & 1.0 & 0.1 & / & / \\
                                  & sigma of Gaussian blurring          & [0.1, 2.0] & [0.1, 2.0] & / & / \\
         \hline
         random solarization      & solarization probability            & 0.0 & 0.2 & / & / \\
         \hline
      \end{tabular}
   }
   \caption{The parameters of data augmentations applied during self-supervised training.
   ``[·, ·]'' denotes the range for uniform sampling.
   }
   \label{table:augmentation}
\end{table*}

\section{Asymmetric Patch Sampling}
To reduce the probability of sampling in the overlapping patch, we adopt an asymmetric sampling strategy, where the sampling can be represented as
\begin{align}\label{eq:sampling}
   p_{\rm sel} = \rho \cdot (1 - r_{\rm overlap})^\gamma,
\end{align}
where $\rho$ is the coefficient required to determined.
So, as shown in Figure~\ref{fig:sampling} of the main paper, the larger $\gamma$ is, the smaller possibility of the sampled overlapping patches is. 
Due to uniform sampling conducted in the first view $x_1^{\prime}$, the probability of patch sampling in the union region $x_1^{\prime} \cap x_2^{\prime}$ can also be regarded as $s_1$. 
In other words, this condition can be presented as
\begin{align}\label{eq:condition}
   \int_{0}^{1} p_{\rm sel} {\rm d}r_{\rm overlap} = s_1.
\end{align}
Combining with Eq.~\ref{eq:sampling}, we obtain
\begin{align}\label{eq:equation}
   & \int_{0}^{1} p_{\rm sel} {\rm d}r_{\rm overlap} = s_1 \nonumber \\
   & \Rightarrow 
   \int_{0}^{1} \rho \cdot (1 - r_{\rm overlap})^\gamma {\rm d}r_{\rm overlap} = s_1 \nonumber \\
   & \Rightarrow 
   \frac{\rho}{\gamma + 1} = s_1 \nonumber \\
   & \Rightarrow \rho = (\gamma + 1) \cdot s_1
\end{align}
Combining with Eq.~\ref{eq:sampling}, we can obtain $p_{\rm sel} = (\gamma + 1) \cdot s_1 \cdot (1 - r_{\rm overlap})^\gamma$, namely Eq.~\ref{eq:sampling} of the main paper.

\section{Algorithm}
To clearly clarify our proposed method, we give the overall algorithm as shown in Algorithm~\ref{alg}.

\renewcommand{\algorithmicrequire}{\textbf{Input:}} 
\renewcommand{\algorithmicensure}{\textbf{Output:}}
\begin{algorithm*}[h]
   \caption{Asymmetric Patch Sampling for Contrastive Learning}
   \label{alg}
   \begin{algorithmic}[1]
      \Require{Unlabeled training data $\mathcal{X} = \{x\}$.}
      \Ensure{The parameters of ViT model $\mathcal{F}(\cdot)$}.
      \State Initialize the parameters of ViT model $\mathcal{F}(\cdot)$, projection head $\mathcal{G}(\cdot)$ and prediction head $\mathcal{H}(\cdot)$;
      \For{number of training iterations}
            \State Obtain augmented views $[ x_1^\prime, x_2^\prime ]$ and crop boxes $[ B_1^\prime, B_2^\prime ]$ by applying augmentation $\mathcal{T}_1(\cdot)$ and $\mathcal{T}_2(\cdot)$, respectively;
            \State Obtain view $x_1$ by uniformly sampling $s \times 100$ percent of the patches from $x_1^\prime$;
            \State Compute patch overlap ratio $r_{\rm overlap}$ between view $x_1$ and view $x_2^{\prime}$;
            \State Sample $s \times 100$ percent of the patches from $x_2^\prime$ to obtain another view $x_2$ by Eq.~\ref{eq:sampling};
            \State Feed $x_1$ and $x_2$ into ViT model $\mathcal{F}(\cdot)$ and projection head $\mathcal{G}(\cdot)$, to obtain $z_1 = \mathcal{G}(\mathcal{F}(x_1))$ and $z_2 = \mathcal{G}(\mathcal{F}(x_2))$;
            \State Feed $z_1$ and $z_2$ into prediction head $\mathcal{H}(\cdot)$, to obtain $q_1 = \mathcal{H}(z_1)$ and $q_2 = \mathcal{H}(z_2)$;
            \State Compute contrastive loss $\mathcal{L}_{\rm contrast}$ cross $z_j$ and $q_k$ by Eq.~\ref{eq:asym-loss}, where $j, k \in \{1,2\}$;
            \State Update the parameters of $\mathcal{F}(\cdot)$, $\mathcal{G}(\cdot)$ and $\mathcal{H}(\cdot)$.

      \EndFor
   \end{algorithmic}
\end{algorithm*}

\section{The Effect of Training Epochs}
\begin{figure*}[ht]
\centering
\subfloat[CIFAR10]{{\includegraphics[width=0.45\linewidth]{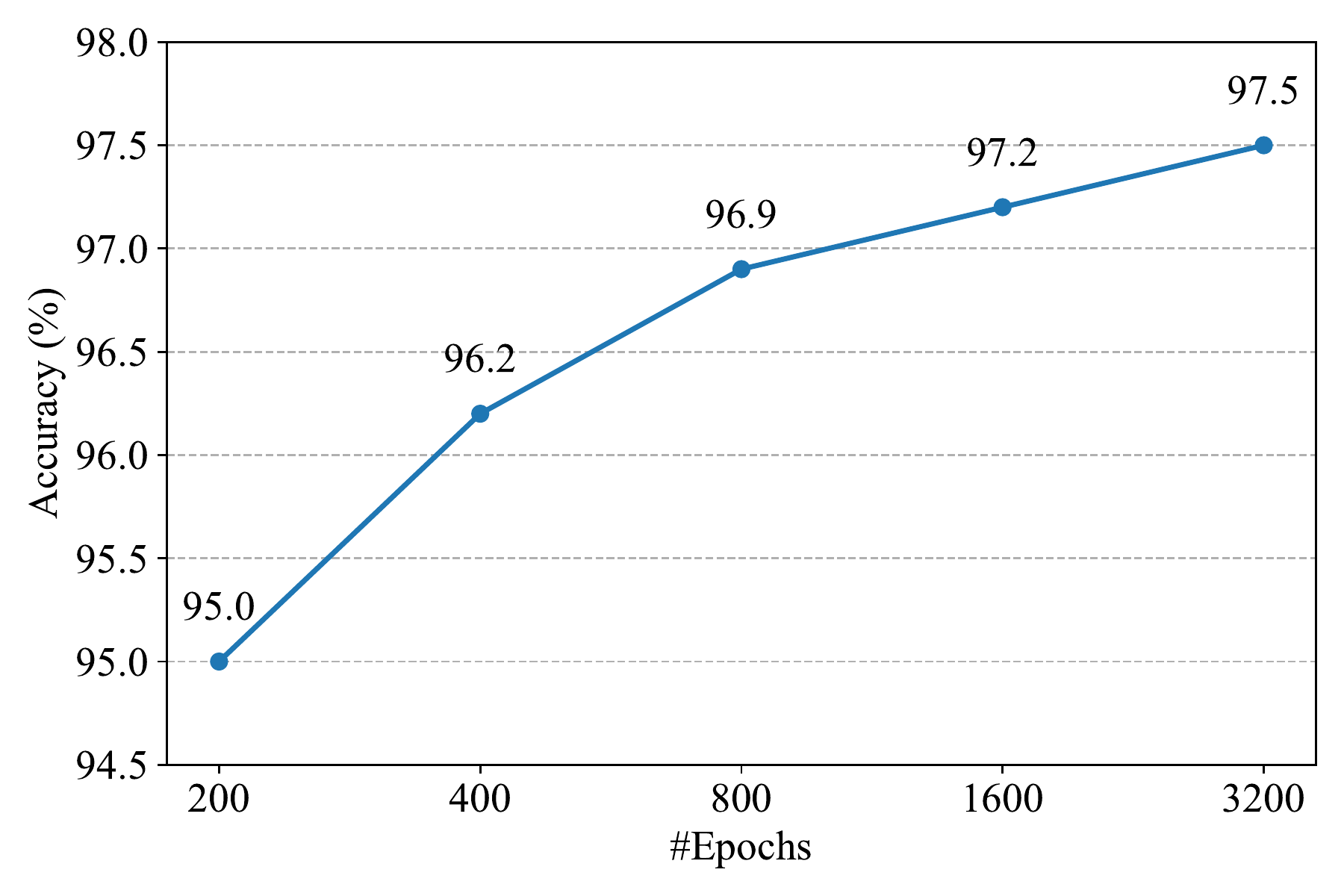}}}
\subfloat[CIFAR100]{{\includegraphics[width=0.45\linewidth]{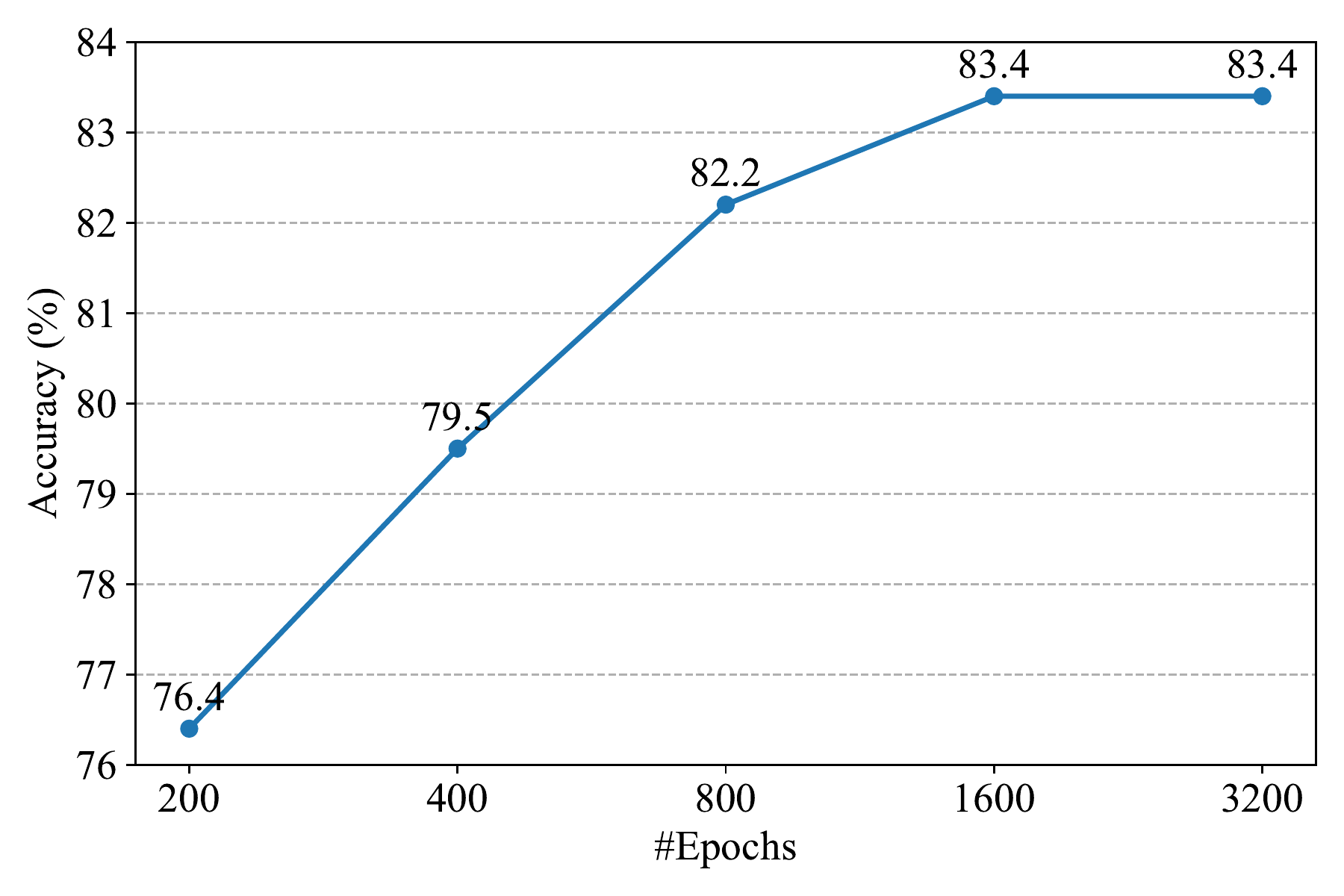}}}
\caption{Training with different numbers of epochs on CIFAR dataset.
}
\label{fig:epochs}
\end{figure*}

To validate the effect of training epochs during self-supervised pretraining, we conduct experiments on ViT-T/2 with different numbers of pretraining epochs.
As shown in Figure~\ref{fig:epochs}, we can find the proposed method achieves better performance with longer training schedule.
Meanwhile, the performance improvement progressively reaches to saturation with the increment of pretraining epochs. 
Hence, we adopt 1600 pretraining epochs in the paper, where the performance is close to saturation.

\end{document}